\documentclass{article} 

\usepackage[ruled, linesnumbered]{algorithm2e}

\newcommand{\ignore}[1]{} 

\usepackage{hyperref}
\usepackage{color}
\usepackage{amsfonts}
\usepackage{amsmath}
\usepackage{amssymb}
\usepackage{float}
\newcommand{\T}[1]{\ensuremath{\mathcal{#1}}} 
\newcommand{\M}[1]{\ensuremath{\mathbf{#1}}} 

\usepackage{amsthm}

\usepackage[usenames,dvipsnames]{xcolor}

\usepackage{tikz}
\usepackage{pgfplots}
\usetikzlibrary{plotmarks}

\usepackage{multirow}

\usepackage{enumitem}

\makeatletter
\newenvironment{customlegend}[1][]{%
	\begingroup
	\pgfplots@init@cleared@structures
	\pgfplotsset{#1}%
}{%
\pgfplots@createlegend
\endgroup
}%

\def\addlegendimage{\pgfplots@addlegendimage}
\makeatother

\definecolor{mycolor1}{rgb}{0.00000,0.44700,0.74100}%
\definecolor{mycolor2}{rgb}{0.85000,0.32500,0.09800}%
\definecolor{mycolor3}{rgb}{0.92900,0.69400,0.12500}%
\definecolor{mycolor4}{rgb}{0.49400,0.18400,0.55600}%
\definecolor{mycolor5}{rgb}{0.46600,0.67400,0.18800}%

\definecolor{mycyan}{rgb}{0.0, 0.75, 1.0}

\newcommand{\nestedsampl}{\textbf{NestedSampling}}
\newcommand{\treeparam}{\textbf{TreeParameterization}}

\usepackage{graphicx}

\newcommand{\MyTikzmark}[2]{%
	\tikz[overlay,remember picture,baseline] \node [anchor=base] (#1) {$#2$};%
}

\newcommand{\DrawVLine}[3][]{%
	\begin{tikzpicture}[overlay,remember picture]
	\draw[shorten <=0.3ex, #1] (#2.north) -- (#3.south);
	\end{tikzpicture}
}

\newcommand{\DrawHLine}[3][]{%
	\begin{tikzpicture}[overlay,remember picture]
	\draw[shorten <=0.2em, #1] (#2.west) -- (#3.east);
	\end{tikzpicture}
}

\usepackage{blkarray}
\newcommand{\matindex}[1]{\mbox{\scriptsize#1}}

\begin{document}

\markboth{I. Perros et al.}{Sparse Hierarchical Tucker Factorization}

\title{Sparse Hierarchical Tucker Factorization and its Application to Healthcare}

\author{Ioakeim Perros, Robert Chen, Richard Vuduc, Jimeng Sun\\ Georgia Institute of Technology}
\date{}
\maketitle
\begin{abstract}
  We propose a new tensor factorization method, called the \emph{Sparse}  \emph{Hierarchical} \emph{Tucker} (Sparse H-Tucker), for sparse and high-order data tensors.
  Sparse H-Tucker is inspired by its namesake, the classical Hierarchical Tucker method, which aims to compute a tree-structured factorization of an input data set that may be readily interpreted by a domain expert.
  However, Sparse H-Tucker uses a nested sampling technique to overcome a key scalability problem in Hierarchical Tucker, which is the creation of an unwieldy intermediate dense core tensor;
  the result of our approach is a faster, more space-efficient, and more accurate method.
  
  We extensively test our method on a real healthcare dataset, which is collected from $30$K patients and results in an 18th order sparse data tensor.
  Unlike competing methods, Sparse H-Tucker can analyze the \emph{full} data set on a single multi-threaded machine.
  It can also do so more accurately and in less time than the state-of-the-art: on a 12th order subset of the input data, Sparse H-Tucker is $18\times$ more accurate and $7.5\times$ faster than a previously state-of-the-art method.
  Even for analyzing low order tensors (e.g., $4$-order), our method requires close to an order of magnitude less time and over two orders of magnitude less memory,
  as compared to traditional tensor factorization methods such as CP and Tucker.
  Moreover, we observe that Sparse H-Tucker scales nearly linearly in the number of non-zero tensor elements.
  The resulting model also provides an interpretable disease hierarchy, which is confirmed by a clinical expert.
\end{abstract}

\section{Introduction} \label{sec:intro}

This paper proposes a new tensor factorization method, designed to model multi-modal data, when the number of modes is high and the input data
are sparse.
Analyzing multi-modal data arises in data mining due to the abundance of information
available that describes the same data objects~\cite{Lahat2014}.
We are motivated to study \emph{tensor methods} because they are recognized as one of the most promising approaches for mining multi-modal data, with proof-of-concept demonstrations in a broad variety of application domains, such as
neuroscience~\cite{Latchoumane2012,Cichocki2013}, epidemics~\cite{Matsubara2014},
human behavior modeling~\cite{Jiang2014},
natural language processing~\cite{Kang2012},
social network analysis~\cite{Papalexakis12}, 
network intrusion detection~\cite{Sun2006}, and healthcare analytics~\cite{Ho2014a,Ho2014b,Wang2015}, to name just a few.
However, tensors also pose numerous computational scalability challenges, in all the senses of time, storage, and accuracy.
This paper addresses these challenges.

By way of background, a tensor generalizes the concept of a matrix to more than two dimensions (rows and columns).
A tensor may be \emph{dense}, meaning one must assume nearly all its entries are non-zero, or \emph{sparse}, meaning most entries are zero, so that tensor may be stored compactly and many computational operations may be eliminated.
In data analysis, each dimension is referred to as a \emph{mode}, \emph{order},
or \emph{way}~\cite{Kolda2009}. For example, a 10th order disease tensor might be constructed
so as to capture interactions across 10 different disease groups.
Examples of well-known tensor decomposition methods include
CP (CANDECOMP-PARAFAC) and Tucker methods~\cite{Harshman1970,carroll1970analysis,tucker1966some,De2000}.
However, despite their known value to data analysis problems, these methods have been largely limited to the analysis of data sets with a relatively small number of modes, typically 3 to 5, and so would not apply to our hypothetical 10th order example.
There are two principal challenges:
\begin{enumerate}[leftmargin=*]
  \item \noindent \emph{Modeling assumptions.}
    Traditional tensor models like CP or Tucker reveal strictly flat structures. By contrast, the 10 different disease groups in our hypothetical example might have natural subgroups, or even hierarchical structure; CP and Tucker ignore the possibility of such structure. Indeed, one might rightfully expect that, as the order grows, so, too, does the number of subgroups or the depth of the hierarchy.

  \item \noindent \emph{Exponential computational cost.}
    With respect to the order of the tensor, there may be exponential costs in space and time.
    In the case of the Tucker method, the cause is the need to store a fully \emph{dense} core tensor $\T{C}$ as output, even if the input tensor is sparse.
    To see why this is problematic, consider an order $d=50$ input tensor for which we wish to compute just a very low-rank approximation of, say, $r=2$.
    Then, the dense core has size $r^d$, which in this case is nearly 9~Petabytes, assuming 8 bytes per (double-precision floating-point) value~\cite{Grasedyck2013}.
\end{enumerate}


To tackle the challenges above, we propose a scalable hierarchical tensor factorization for sparse high-order tensors, which we call the \emph{Sparse Hierarchical Tucker} (or Sparse H-Tucker) method.
Sparse H-Tucker expresses
mode interactions as a binary tree,
which is further parameterized in order to allow the approximation accuracy and cost to be tuned.
For the same approximation error, it provides \textit{close to an order of magnitude gain}
in terms of the \textit{time} required, when compared to a state-of-the-art CP factorization,
and \textit{over two orders of magnitude gain}
in terms of the \textit{space} required, when compared to a state-of-the-art Tucker factorization method.
At the same time, it respects \emph{sparsity} in the input, achieving a \textit{near-linear scale-up} in time and space
with respect to the non-zeros of the input
tensor.
Perhaps somewhat surprisingly, this level of performance
is \emph{not} achieved at the cost of accuracy; on the contrary, as we verify experimentally,
Sparse H-Tucker achieves
remarkable gains in accuracy as well, particularly as the tensor density and order increase.

\begin{figure}
\centering
\includegraphics[scale=.25]{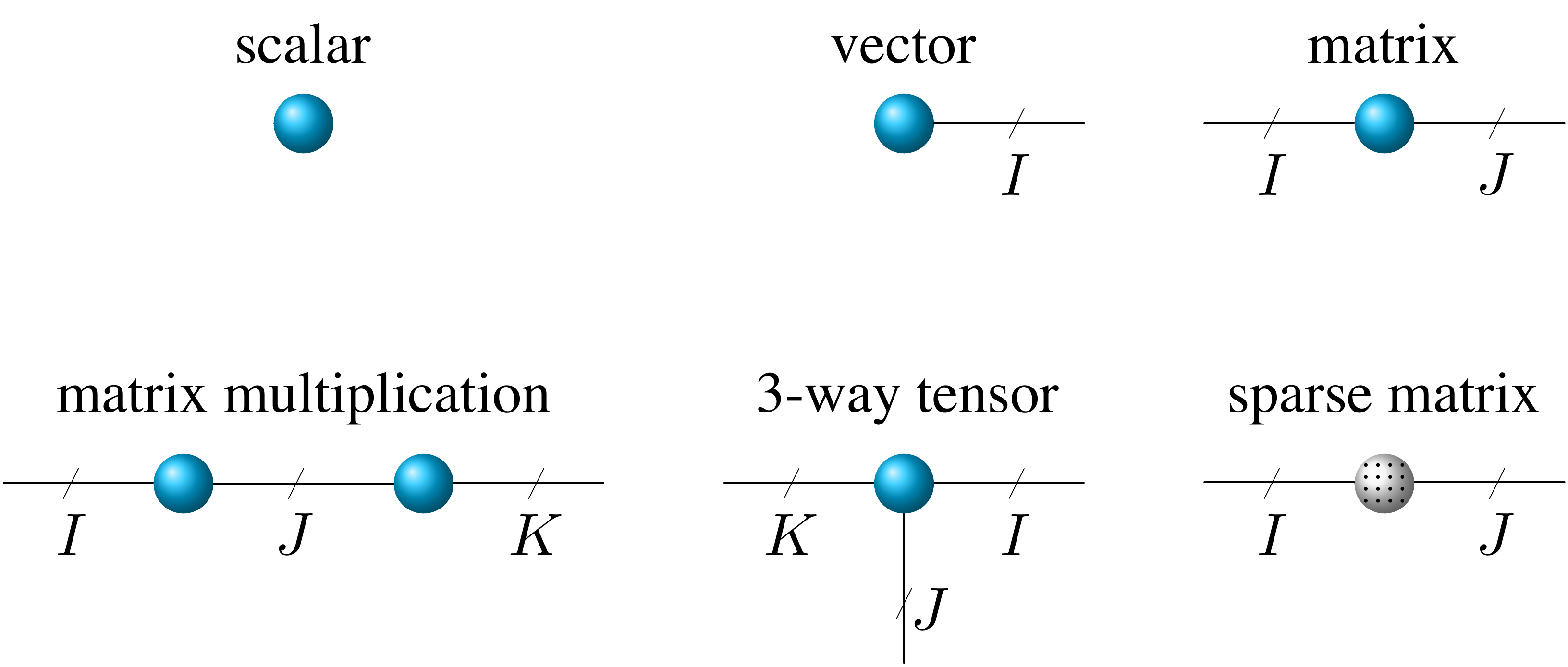}
\caption{Basic tensor network notation.}
\label{fig:TN}
\end{figure}

Another subtle but important challenge in dealing with high-order tensors is the lack of \textit{intuitive}
and \textit{generic representation} for tensors and tensor 
operations, which may hinder end-user analysts from adopting tensor methods.
As a result, most works on new tensor models are presented for a specific low-order tensor (e.g., 3 orders~\cite{rendle2010pairwise,fang2014fast}).
For example, recent work by Fang et al.~models the interactions of each one of the three modes
with the two others through a tensor, the horizontal slices of which
are further decomposed into two low-rank matrices~\cite{fang2014fast}.
The case of $d > 3$, when each horizontal slice would be a tensor,
is not addressed. In order to tackle this limitation as well,
we adopt a recently proposed tensor formalism
called {\it tensor networks}, originally developed for applications in quantum chemistry and physics~\cite{Chinnamsetty2007,Kazeev2014,Khoromskij2007}.
This formalism has a nice visual representation as well, the basic elements of which appear in Figure~\ref{fig:TN} and are reviewed in Section~\ref{sec:Backg}.

Besides their simple and intuitive graphical representations, 
tensor networks also provide a set of computational strategies 
to approximate a high-order dense tensor by an interconnected graph of low-order tensors
(typically, 2nd and 3rd order tensors)~\cite{Oseledets2011,Grasedyck2010,Orus2014a,Cichocki2014a}.
These methods enable the compression of a tensor of size $n^d$
into a form that is linear in $d$, while preserving favorable numerical
properties. 
However, successfully applying tensor networks to unsupervised learning has not been demonstrated in practice.
One reason is that, despite their nice theoretical properties, tensor network methods target
\textit{dense} tensors, which is the usual case in quantum chemistry and physics applications;
by contrast, data tensors are usually \textit{sparse}.
Also, the design of tensor networks has hereto focused
on compression, rather than interpretation and pattern discovery, though their potential for the latter has been appreciated by some.\footnote{The intuition behind and potential applications of tensor networks in data processing appear in recent surveys by Cichocki~\cite{Cichocki2014a,Cichocki2014b}.}
Nevertheless, to our knowledge, this paper is the first to try to really \emph{apply} tensor network modeling to a knowledge discovery application,
through experimental evaluation as well as discussion on the interpretability of the model.

\ignore{Our contributions can be summarized as follows:
\begin{itemize}
	
	\item {\bf Tensor networks for data mining:} We \emph{apply}
	 the tensor networks formalism 
	 as a modeling tool for data mining, and evaluate its practicality.
	
	\item {\bf A sparse high-order tensor factorization method:}
	We propose {\it Sparse H-Tucker}, a scalable tensor factorization method tailored
	to sparse and high-order tensors, which enjoys
	a near-linear scale-up to the number of non-zero tensor elements in
	both time and space. The method provides orders of magnitude
	improvements in cost-approximation quality trade-offs
	as compared to standard tensor factorization methods such as Tucker and CP.
	
	\item {\bf Demonstration on a real-world healthcare application:} We present a case study in which we apply
	  Sparse H-Tucker to disease phenotyping using electronic health records (EHR).
          The interpretability of the resulting hierarchical disease model
	  is confirmed by a domain expert.
\end{itemize}
}
An earlier version of the present work appeared in the proceedings of IEEE ICDM 2015~\cite{perros2015}.
In addition to the ones of~\cite{perros2015}, our contributions in this extended version can be summarized as follows:
\begin{itemize}
	
	\item {\bf Tensor networks for data mining:} We showcase how our method may be derived and understood
	from the vantage point of the \emph{tensor networks} mathematical formalism, rather
	than strictly algebraic representations, which we review. This description renders 
	the work simpler and more easily accessible. 
	This work is the first to \emph{apply}
	the tensor networks formalism as a modeling tool for unsupervised learning purposes, and evaluate its practicality.
	
	\item {\bf Theoretical backing and explanations:}
	We underpin {\it Sparse H-Tucker} with the necessary theoretical proofs and extensive discussions
	required to fully understand its mathematical foundations and its functionality.
	
	\item {\bf Thorough experimental evaluation on real healthcare data:} We complement our
	disease phenotyping case study using electronic health records (EHR), by
	providing experimental evaluation of the low-order scenario as well; in that case,
	we identify \textit{close to an order of magnitude gains in time} and \textit{over two orders of magnitude gains
	in memory}, as compared to traditional tensor factorization methods. As such, we
	justify the suitability of our work to low-order tensor problems as well.
	We also provide a detailed discussion on the interpretability of the resulting hierarchical disease model,
	guided by a domain expert.
\end{itemize}

\ignore{The rest of the paper is organized in the following way:
Section~\ref{sec:Backg} contains the necessary background.
Section~\ref{sec:shtucker} describes Sparse H-Tucker, the proposed model
and the factorization algorithm we propose to achieve it.
Section~\ref{sec:Experiments} presents experimental evaluation with real healthcare
data. Finally, we conclude this work in Section~\ref{sec:conclusion}.}

\section{Background} \label{sec:Backg}

This section introduces the necessary definitions and
 the preliminaries of matrix and tensor operations. Table~\ref{table:notation} lists the notations used throughout the paper.

\subsection{Matrix factorizations}
The Eckhart-Young Theorem for the \textit{Singular Value Decomposition}
(SVD)~\cite{Golub2012} for $\M{U} \M{\Sigma} \M{V} = svd(\M{A})$, where
$\M{A} \in \mathbb{R}^{m \times n}$ defines
that if $ k < r = rank(\M{A}) $ and
$\M{A}_k  = \sum_{i=1}^k \sigma_i u_i v^T_i $, then: $\underset{rank( \M{B}) = k}{min} || \M{A} - \M{B} ||_2 = || \M{A}  - \M{A_k} ||_2 = \sigma_{k+1} $.
Instead of using the singular vectors in SVD, the
\textit{CUR decomposition}~\cite{Mahoney2009} uses representative
columns $\M{C}$ and rows $\M{R}$ to approximate the input matrix.
The relative-error guarantees of this method~\cite{Drineas2008} depend on the
notion of \textit{leverage score sampling}\footnote{Alternative ways of 
lower-cost sampling (e.g. based on the row/column norms) are known
to give much coarser additive error estimates~\cite{Drineas2006}.}.
The leverage scores $\pi$ for each $j=1, \dots, n$ column
of $\M{A}$ are: $\pi_j = 1/k \sum_{\xi = 1}^k \left( v_j(\xi) \right)^2 $,
where $v_j$ is the $j$-th right singular vector of $\M{A}$
(out of $k$ computed in total).
Symmetrically, row sampling for matrix $\M{R}$ is achieved by
applying the above process on $\M{A}^T$.
It is proven that sampling $\mathcal{O}( k logk / \epsilon^2)$ columns (and rows of $\M{A}^T$)
based on the distribution $\pi$ and defining $\M{U} = \M{C}^{+} \M{A} \M{R}^{+}$ gives:
$ || \M{A} - \M{C}\M{U}\M{R} ||_F \leq (2 + \epsilon) || \M{A} - \M{A_k} ||_F $. 
Tensor versions of CUR approximation are given in~\cite{Mahoney2006,Tsourakakis2010},
but cannot handle high-order tensors due to cost limitations
faced, similar to the ones of Tucker.

\subsection{Tensor operations and	factorizations}
Tensors are high-order generalizations of matrices.
A \textit{fiber} is a vector extracted from a tensor by
fixing all modes but one and a slice is a matrix extracted from a tensor by
fixing all modes but two~\cite{Kolda2009}. Let a $d$-order tensor $\T{A}\in\mathbb{R}^I$.
The index set over which the
tensor is defined is: $ I := I_1 \times \dots \times I_d $ and the
index set of each individual mode is $ I_{\mu} :=
 \{ 1, \dots, n_{\mu} \}, \mu \in \{ 1, \dots, d \} $.

\textit{Matricization (or reshaping, unfolding)} logically reorganizes tensors
into other forms, without changing the values themselves.
Let the index set $ I^{(\mu)} := I_1 \times \dots \times I_{\mu-1} \times I_{\mu+1} \times
\dots \times I_d  $. Then, the \textit{$\mu$-mode matricization} is a mapping:
$ \M{A}^{(\mu)}: \mathbb{R}^I \rightarrow  \mathbb{R}^{I_{\mu} \times I^{(\mu)} } $.
As a result, the mode-$\mu$ fibers of the tensor become columns of a matrix.
Given $\M{U}_{\mu} \in \mathbb{R}^{J_{\mu}
\times I_{\mu} } $, the \textit{$\mu$-mode multiplication} is defined
by $ ( \M{U}_{\mu} \circ_{\mu} \T{A} )^{(\mu)} := \M{U}_{\mu} \M{A}^{(\mu)} \in
\mathbb{R}^{ J_{\mu} \times I^{(\mu)} } $. Given matrices 
$\M{U}_v \in \mathbb{R}^{J_v \times
I_v}$ with $v = 1, \dots, d$ the \textit{multi-linear multiplication} is defined as:
$  (\M{U}_1, \dots, \M{U}_d) \circ \T{A} := \M{U}_1 \circ_1 \dots
\M{U}_d \circ_d \T{A}  \in \mathbb{R}^{J_1 \times \dots \times J_d} $.



The factorization of a tensor into a sum
of component rank-one tensors
is called the \textit{CP/PARAFAC}~\cite{Harshman1970,carroll1970analysis} factorization.
If the rank of a $d$-order tensor $\T{A}$ is equal to $R$, then:
$ \T{A} = \sum_{r=1}^R \lambda_r ~ a_r^{(1)} \bullet a_r^{(2)} \bullet \dots \bullet a_r^{(d)} $.
The most popular factorization method approximating the above model
is the CP-Alternating Least Squares (ALS)~\cite{Harshman1970,carroll1970analysis,Kolda2009}, which
optimizes iteratively over each one of the output matrices by fixing all others.
The \textit{Tucker} format is given by the following form~\cite{tucker1966some,De2000}:
$ \T{A} = (\M{U}_1, \dots, \M{U}_d)  \circ \T{C}  $,
where  $\M{U_{\mu} \in \mathbb{R}^{n_{\mu} \times k_{\mu}}}$ are (columnwise) orthonormal matrices
and  $\T{C} \in \mathbb{R}^{k_1 \times \dots \times k_d}$ is a core tensor,
The tuple $(k_1, \dots, k_d)$ with (elementwise) minimal entries
for which the above relation holds is called the Tucker rank.
In data analysis applications, the above relation is expected
to hold only approximately. For fixed $\M{U}_{\mu}$
matrices, the unique core tensor minimizing the approximation error
is given by: $\T{C} = (\M{U}^T_1, \dots, \M{U}^T_d)  \circ \T{A} $.
If the core tensor is computed in the above way and
each $\M{U}_{\mu}$ contains
the leading $k_{\mu}$ left singular vectors
of $ \T{A}^{(\mu)} $,
the factorization of tensor $\T{A}$
is called the higher-order SVD (HOSVD)~\cite{De2000,Kolda2009}.
HOSVD is considered as a good initialization to the
higher-order orthogonal iteration (HOOI)~\cite{de2000best},
which is also an ALS-type algorithm, being the most popular
way to approximate the Tucker format in real world applications.



\subsection{Tensor networks}
A \emph{tensor network diagram}, or just \emph{tensor network} hereafter, provides an
intuitive and concise graphical notation for representing tensors and
operations on tensors~\cite{Cichocki2014a,Cichocki2014b}.
A scalar, vector, matrix, or tensor is represented by the ``ball-and-stick" symbol
that appears in Figure~\ref{fig:TN}, where each circle denotes the object and each edge an order or mode.
Annotated circles indicate special structure, such as being sparse.
Where an open edge represents a mode, a closed edge that connects two tensors
represents a contraction along the given edge.
Contracting two tensors $\T{A}\in \mathcal{R}^{I_1\times\dots\times I_N}$ and
$\T{B} \in \mathcal{R}^{J_1\times \dots \times J_M}$ on common modes
$I_n=J_m$ yields another tensor, $\T{C}\in \mathcal{R}^{I_1\times\dots\times I_{n-1}\times I_{n+1}\times \dots\times I_N \times J_1\times \dots \times J_{m-1}\times J_{m+1}\times\dots\times J_M}$.

\noindent\textbf{Hierarchical Tucker and its Limitations}
One popular model of the tensor network family
that shares structural similarities with our proposed model
is the Hierarchical Tucker (H-Tucker in short) 
presented in~\cite{Grasedyck2010}.
 Intuitively, the H-Tucker factorization algorithm proposed in~\cite{Grasedyck2010}
 first decomposes the input tensor into the Tucker format through the
 HOSVD and then recursively factorizes the output tensor of this process.
 Such a strategy though suffers from severe scalability issues as the tensor order $d$ increases.
 Despite the fact that the final form of H-Tucker
 requires linear storage to $d$, 
 the size of the intermediate core tensor computed
 increases \textit{exponentially} to $d$;
 and this core tensor is \textit{dense}.
 As a result, this method faces
 a potential \textit{memory blow-up} as it requires further decomposing an intermediate result that may not even fit into memory. 
 
%
 

Another factorization scheme that is based on H-Tucker
and is similar to ours was proposed in the tensor community by Ballani et
al~\cite{Ballani2013,Ballani2012}.  However, that work exclusively targets 
dense tensors (does not work for sparse input),
while ours focuses on sparse ones and data mining applications.

\begin{table}
	\centering
	\begin{tabular}{|c|c|}
		\hline
		Symbol & Description \\
		\hline \hline 
		$\T{A}$      &  tensor A        \\     \hline
		$\M{A} $     & matrix A  \\ \hline
		$\M{A}^{+}$ & pseudo-inverse of A \\ \hline
		$\otimes$     &  Kronecker product         \\ \hline
		$\bullet$ & vector outer product \\ \hline
		$<a, b> $ & vector inner product \\ \hline
		$\mathcal{T_I}$ & dimension tree \\ \hline
		$\mathcal{L(T_I)}$ & leaf tree nodes \\ \hline
		$\mathcal{I(T_I)}$ & interior tree nodes \\ \hline
		$s(t) $ & set of successors of parent node $t$ \\ \hline
		$t_r$ & root node of the dimension tree \\ \hline
		$ I_t $ & index set of subset $t$ of modes \\ \hline
		$| I_t | $ & cardinality of set $I_t$ \\ \hline
	\end{tabular}
	\caption{List of notations used}
	\label{table:notation}
\end{table}
\section{Sparse Hierarchical Tucker} \label{sec:shtucker}


\subsection{Model}\label{sec:Htuckermodel}
\begin{figure}
	\centering
	\includegraphics[scale=0.25]{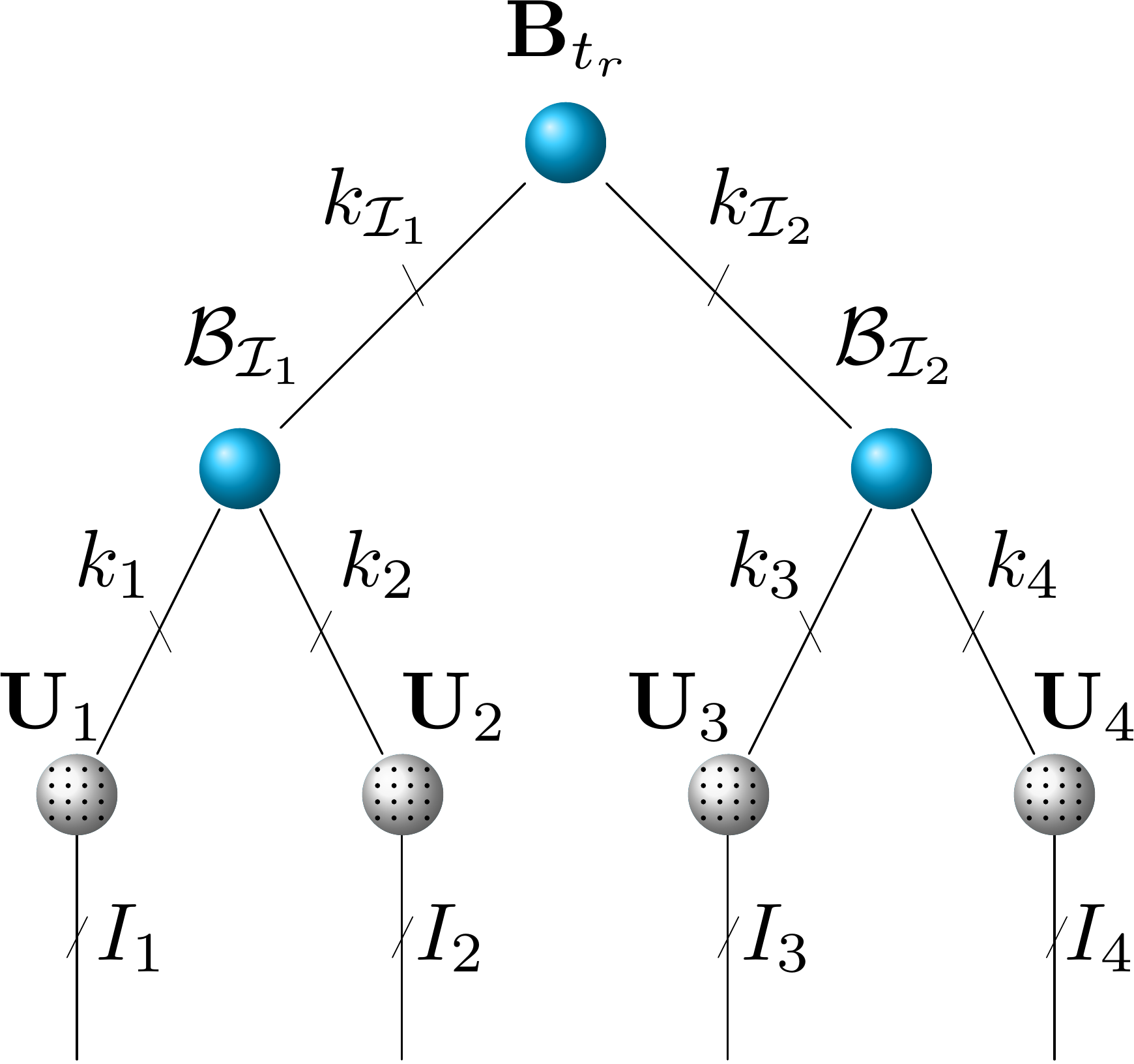}
	\caption{Sparse Hierarchical Tucker format in tensor
		network notation (4-order tensor, balanced dimension tree).}
	\label{fig:ht_sparse}
\end{figure}

Our proposed target model is called
the \textit{Sparse Hierarchical Tucker}
(Sparse H-Tucker).
An example of this model in tensor network notation appears in Figure~\ref{fig:ht_sparse}.
In Sparse H-Tucker, the tensor modes are split recursively, resulting in a binary tree that we call the \textit{dimension tree} and denote by $\mathcal{T_I}$. Each node
of this tree contains a subset $t \subset \{1, \dots,
d \}$ of the modes and is either a leaf and singleton
$t={\mu}$ or the union of its two disjoint successors $t_1, t_2: t = t_1 \cup t_2 $.
Each tree node is associated with
an output factor of the model.
We denote these output factors by,
\[ (\T{B}_t)_{t \in \mathcal{I}(T_I)} \in
\mathbb{R}^{k_t \times	k_{t_1} \times k_{t_2}}, (\M{U}_{t})_{t \in \mathcal{L}(T_I)}
\in \mathbb{R}^{I_t \times k_t}. \]
The tensors
$\T{B}_t$ are called the \textit{transfer tensors}, which
correspond to the interior nodes, $\mathcal{I(T_I)}$;
the matrices $\M{U}_t$ correspond
to the leaves of the tree, $\mathcal{L(T_I)}$, where
$s(t) = \{t_1, t_2\}$ denotes the set of successors of node $t$.
By definition, the matrices $\M{U}_t$  associated with
the leaves of this tree structure are sparse.
The tensor associated with the root node $t_r$
is a degenerate one (i.e., it is a matrix since $k_{t_r}=1$), because unlike other interior nodes, only the root node
connects to 2 nodes instead of 3.


Our proposed model's tree structure is like that of the H-Tucker model~\cite{Grasedyck2010}.
However, Sparse H-Tucker preserves sparsity. 
By contrast, in H-Tucker, the matrices corresponding to the leaf nodes are dense, which fundamentally limits the scalability of any algorithms operating on it.


\subsection{Sparse H-Tucker factorization algorithm}

The proposed factorization method can be conceptually
divided into two phases:
\begin{itemize}[leftmargin=*]
\item \noindent{\bf Phase 1} computes a \textit{sampling-based} low-rank approximation
of all $\M{A}^{(t)}$ associated with each tree node
except for the root. Notice that $\M{A}^{(t)}$ combines all modes contained in
$t$ as row indices and the rest of the modes into column indices.
\item \noindent{\bf Phase 2} uses the output of Phase 1,
in order to assemble the final Sparse H-Tucker model in parallel.
\end{itemize}
The rationale behind these two phases is first to conduct all the preparation work in Phase 1 and then to compute the 
expensive steps fully in parallel in Phase 2. 

\begin{algorithm}
			\KwData{Input tensor $\T{A} \in \mathbb{R}^I$, tree
			$\mathcal{T_I}$, accuracy parameter $\epsilon$}
			\KwResult{$(\T{B}_t)_{t \in \mathcal{I(T_I)}}, (\M{U}_{t})_{t \in \mathcal{L(T_I)}}$}
			\tcp{Phase 1}
			$\{P_t, Q_t, \M{M_t}\}= \treeparam(\T{A}, t_r, \varnothing, \epsilon)$ 
			\label{alg:shtucker:line1} \;
			\tcp{Phase 2: fully-parallelizable loop}
			\ForEach{$t \in \mathcal{T_I}$}{
			\uIf{$t  \in \mathcal{I(T_I)}$} {
			Compute $\T{B}_t$ through Equation~\ref{eq:B_constr} \;
		}\Else{ \tcp{$t \in \mathcal{L(T_I)}$}
			Compute sparse matrix $\M{U}_t$ through Equation~\ref{eq:U_constr} \;
			}
		}
	\caption{Sparse Hierarchical Tucker factorization}
	\label{alg:shtucker}
\end{algorithm}
\begin{algorithm}
			\KwData{Tensor $\T{A},$ tree node $t,$ sampled column indices $Q_t,$ accuracy parameter $\epsilon$}
			\KwResult{$\{ P_t, Q_t, \M{M_t} \} \forall t \in \mathcal{T_I}  \backslash t_r$} 
			$\{ t_1, t_2 \} = s(t)$ \;
			$[P_{t_1}, Q_{t_1}, \M{M}_{t_1}, \T{A}_1 ] = \nestedsampl( \T{A}, t_1, Q_t,\epsilon )$
			\label{alg:shtucker2:line3}	\;
			$[P_{t_2}, Q_{t_2}, \M{M}_{t_2}, \T{A}_2 ] = \nestedsampl( \T{A}, t_2, Q_t, \epsilon )$
			\label{alg:shtucker2:line4} \;
			\If{$t_1 \in \mathcal{I(T_I)}$  } {
			$\treeparam(\T{A}_1, t_1, Q_{t_1}, \epsilon)$
		}
			\If{$t_2 \in \mathcal{I(T_I)}$  }{ 
			$\treeparam(\T{A}_2, t_2, Q_{t_2}, \epsilon)$
	}
	\caption{{\small \treeparam}}
	\label{alg:shtucker2}
\end{algorithm}

Algorithm~\ref{alg:shtucker} is our top-level
procedure to compute the Sparse H-Tucker form.
It takes as input the original tensor $\T{A}$, the dimension tree structure
$\mathcal{T_I}$ and a parameter $\epsilon$ which governs the accuracy
of low-rank approximations.
In Line~\ref{alg:shtucker:line1}
of Algorithm~\ref{alg:shtucker}, we invoke  Algorithm~\ref{alg:shtucker2}, by starting the recursion from the root node
of the tree ($t_r$) to parameterize the dimension tree.

Within Algorithm~\ref{alg:shtucker2},
Lines~\ref{alg:shtucker2:line3} and \ref{alg:shtucker2:line4} call
the function  \nestedsampl~to compute the factors
for the approximation of each $\M{A}^{(t)}$.
If $\M{C}_t$ and $\M{R}_t$ contain
column and row samples from $\M{A}^{(t)}$,
respectively, and $\M{M}_t $ is a small matrix minimizing the error of approximation, then the
the product $\M{C}_t \M{M}_t \M{R}_t$
is an approximation of   $\M{A}^{(t)}$.
To avoid the materialization
of $\M{C}_t$ and $\M{R}_t$, we  maintain the index sets $P_t, Q_t$
denoting the row and column indices sampled
from  $\M{A}^{(t)}$ respectively.
The challenges emerging so as to execute the \nestedsampl~function
and its exact operation will be explained in Section~\ref{sec:nestedsampling}.
The recursive procedure \treeparam~is
continued until we reach the leaf nodes.
\footnote{A remark regarding Algorithm~\ref{alg:shtucker2} is that only for the root
node's successors (i.e.,~when $ \{t_1,
t_2 \} = s(t_r) $), it holds that: $ {\T{A}^{(t_1 )}}^T  = \T{A}^{( t_2 )} $.
To reduce redundant computations
within the actual implementation, Line~\ref{alg:shtucker2:line4} of Algorithm~\ref{alg:shtucker2}
is executed only in the case when $t \neq t_r$. Otherwise~($t=t_r$), we set:
$P_{t_2} = Q_{t_1}, Q_{t_2} = P_{t_1}, \M{M}_{t_2} =  \M{M}^T_{t_1}$.}

In Phase 2 of Algorithm~\ref{alg:shtucker},
we construct the output factors of the Sparse H-Tucker model,
by exploiting the  sampling results from Phase 1.
Since the construction over a single node is \textit{completely independent} to others, we can fully parallelize this step.

To assemble the matrices  $\M{U}_t$ corresponding to the leaf
nodes, we directly sample from the column fibers of $\M{A}^{(t)}$:
\begin{equation}
( (\M{U}_t)_i )_{t \in  \mathcal{L(T_I)} } = \M{A}^{(t)}(:, q_i) , q_i \in Q_t.
\label{eq:U_constr}
\end{equation}
Since we are sampling directly from the sparse input tensor for
the construction of the $ (\M{U}_t)_{t \in  \mathcal{L(T_I)} }$ matrices,
our leaf output factors \textit{maintain the sparsity of the input tensor}.
Thus, the requirement of our model for
sparsity on matrices associated with leaf nodes is satisfied.

A great advantage of the model
is that the transfer tensors
are directly assembled
without the need of computing a huge, dense 
intermediate result (as in the case of the
H-Tucker model).
Below, we provide the equation for computing the 
factors $(\T{B}_t)_{t \in \mathcal{I(T_I)}}$ for the interior tree nodes.
The proof of
its correctness is given in the Appendix.
Given nodes $t, t_1, t_2$ where $ \{ t_1, t_2 \} = s(t) $:
\begin{equation}
(\T{B}_t)_{i, j, l} = \sum_{p \in P_{t_1}} \sum_{q \in P_{t_2}}
(\M{M}_{t_1})_{q_j, p} \M{A}^{(t)}_{(p,q), q_i} (\M{M}_{t_2})_{q_l, q},
\label{eq:B_constr}
\end{equation}
where $q_i \in Q_t, q_j \in Q_{t_1}, q_l \in Q_{t_2}$.


\subsection{Tensor approximation via the model's factors} 
Below, we describe how to  approximate the input tensor through the Sparse H-Tucker model.
First, each pair of leaves (matrices) that share a parent (tensor)
are combined into 
a matrix $\M{U}_t$ as follows:
\begin{equation}
(\M{U}_t)_i = \sum_{j=1}^{k_{t_1}} \sum_{l=1}^{k_{t_2}}
(\T{B}_t)_{i, j , l} \left( (\M{U}_{t_1})_j \otimes (\M{U}_{t_2})_l \right)
\label{eq:nestedness}
\end{equation}
where $\{t_1, t_2\} = s(t)$, 
$\T{B}_t \in \mathbb{R}^{k_t \times k_{t_1} \times k_{t_2}}$,
$\M{U}_t \in \mathbb{R}^{(I_{t1}I_{t2}) \times k_t}$,
$\M{U}_{t_1} \in \mathbb{R}^{I_{t_1} \times k_{t_1}}$,
and
$\M{U}_{t_2} \in \mathbb{R}^{I_{t_2} \times k_{t_2}}$.
This process is followed for all interior nodes in a bottom-up fashion.



Given that we have re-constructed the matrices $\M{U}_{t_1},
\M{U}_{t_2}$ ($ \{t_1, t_2\} = s( t_r )$), corresponding to the second level
of the tree, the final input tensor approximation is given in
vectorized form as follows:
\begin{equation}
vec(A) \approx \sum_{j=1}^{k_{t_1}} \sum_{l=1}^{k_{t_2}}
(\M{B}_{t_r})_{j , l} \left( (\M{U}_{t_1})_j \otimes (\M{U}_{t_2})_l \right)
\label{eq:reconstruct}
\end{equation}
Equation~\ref{eq:reconstruct} is a special case of Equation~\ref{eq:nestedness},
accounting for the root node being associated with a matrix rather than a tensor.

We need not construct the full representation if we need only specific
reconstructed entries, such as the reconstruction of a tensor's sub-block.
Instead, we just have to prune the $\M{U}_t$ matrices
associated with the leaves, so that each one
only contains the rows
corresponding to the desired mode indices.
A special case of this property is an element-wise query,
when out of each $\M{U}_t$ leaf matrix we use a
single row vector for the desired element's approximation.
For example, the reconstruction of $\T{A}(i,j,k)$ cell
of a $3$-order tensor $\T{A}$ requires the $\M{U}_{t_1}(i,:),
\M{U}_{t_2}(j,:), \M{U}_{t_3}(k,:)$ to be used as input,
if $t_1,t_2,t_3$ correspond to the mode sets of the leaves.

The equations that govern the reconstruction of our model
also apply in the H-Tucker model~\cite{Grasedyck2010},
where Equation~\ref{eq:nestedness} reflects a property called
\textit{nestedness}; we will use the same terminology hereafter.

\subsection{Nested sampling} \label{sec:nestedsampling}
Below, we describe the \nestedsampl~function that is
called within Algorithm~\ref{alg:shtucker2}. Its role
is to compute the factors required to approximate the
matricizations $\M{A}^{(t)}$ for each subset of modes $t$
associated with each tree node.  Our approach is to form the factors approximating
$\M{A}^{(t)}$ through the CUR decomposition based on leverage score
sampling~\cite{Mahoney2009}. The biased samples from CUR decomposition help
to boost the accuracy of our approach. 
 More specifically, we follow the same sampling strategy as in~\cite{Mahoney2009},
by retrieving $\mathcal{O}( k \log k / \epsilon^2)$ rows or columns
for each required approximation, where $k$ is the rank of SVD,
which is set to a small value ($k=5$)\footnote{We detected no significant change
in the accuracy of the final approximation by tuning $k$, hence we keep it fixed.}. 

However, a simple application of the CUR decomposition within
our factorization framework would completely fail,
due to challenges related to the \textbf{consistency}
of each $\M{A}^{(t)}$ approximation with the whole framework. 
 Assume calling the \nestedsampl~function with arguments $\T{A}, t_1, Q_t,\epsilon$ (as happens in
Line~\ref{alg:shtucker2:line3} of Algorithm~\ref{alg:shtucker2}). Before
even attempting to execute a nested sampling
of $\M{A}^{(t_1)}$, we have to ensure that the available set of
rows and columns is \textit{consistent} across the entire dimension tree.
In other words, we have to ensure that the way we extract
our model's transfer tensors (Equation~\ref{eq:B_constr})
is consistent 
to each individual $\M{A}^{(t)}$ approximation.

To do so, we have to guarantee the validity
of the \textit{nestedness property} (Equation~\ref{eq:nestedness}).
The way we exploit this property towards the proof of correctness of
Equation~\ref{eq:B_constr} is contained in the Appendix.
In the following, we will explain the manner in which we guarantee that
this property holds and how this relates to the column indices $Q_t$  in each \nestedsampl~call
and tensors $\T{A}_1, \T{A}_2$ in each \treeparam~call
of Algorithm~\ref{alg:shtucker2}.

\begin{figure}
	\centering
	\includegraphics[scale=.3]{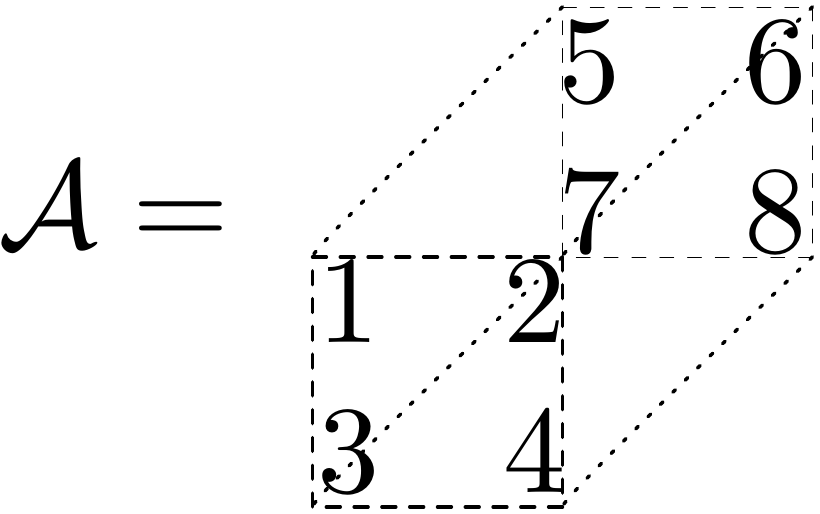} $\M{A}^{(t)} = \begin{blockarray}{cc}
	\textcolor{red}{\mathit{1}} & \mathit{2}\\
	\begin{block}{[cc]}
	1 & 5 \\
	3 & 7 \\
	2 & 6 \\
	4 & 8 \\
	\end{block}
	\end{blockarray}$
	\[
	\M{A}^{(t_1)} = \begin{blockarray}{cccc}
	\mathit{(1,\textcolor{red}{1})} &  \mathit{(2,\textcolor{red}{1})} &
	\mathit{(1,2)} & \mathit{(2,2)} \\
	\begin{block}{[cccc]}
	1 & 2 &  5 & 6 \\
	3 & 4 &  7 & 8 \\
	\end{block}
	\end{blockarray} \] \[  \M{A}^{(t_2)} = \begin{blockarray}{cccc}
	\mathit{(1,\textcolor{red}{1})} &  \mathit{(2,\textcolor{red}{1})}&
	\mathit{(1,2)} & \mathit{(2,2)} \\
	\begin{block}{[cccc]}
	1 & 3 &  5 & 7 \\
	2 & 4 &  6 & 8 \\
	\end{block}
	\end{blockarray} \]
	\caption{Nested index restriction within the sampling framework.
		Let $\{t_1, t_2\} = s(t), t = \{1,2\}, t_1 = 1, t_2 = 2,
		Q_t = \{1\}, I_{t_2}=\{1,2\}$.
		Then, the successor node associated with $t_1$
		can sample (and fill $Q_{t_1}$) from one of the two first columns of $\M{A}^{(1)}$, since $Q_t$ is restricted.}
	\label{fig:nestedness2}
\end{figure}

Equation~\ref{eq:nestedness} dictates
that we should be able to construct
each column vector of $(\M{U}_{t})_{t \in \mathcal{I(T_I)}}$ through
linearly combined Kronecker products of column vectors of
$\M{U}_{t_1}, \M{U}_{t_2}$.
Within our framework, where the $\M{U}_{t}$ matrices
contain \textit{actual fiber samples} from the input tensor,
this restriction is translated to enforcing the following
reduction on the available column fibers for the CUR decomposition:
\begin{equation}
Q_{t_1} \subseteq I_{t_2} \times Q_{t}
\label{eq:restrict}
\end{equation}
where $\{t_1, t_2\} = s(t)$.  The notation $\times$ denotes the cartesian set product.
Relation~\ref{eq:restrict} implies that the columns sampled ($Q_{t_1}$) by the successor node associated with $t_1$ will be a subset
of all possible combinations formed between the
unrestricted index set $I_{t_2}$ and the fixed index set $Q_t$
(which had been previously defined  by the parent node).
By symmetry, $Q_{t_2} \subseteq I_{t_1} \times Q_t $.
In order to clarify this index restriction, we use a toy
example in Figure~\ref{fig:nestedness2}. Given
that the $1$st column of $\M{A}^{(t)}$ is selected
(which means that $Q_t = \{1\}$),
the available fibers for the successor node
are those containing the $1$st element of the $3$rd mode in their
multi-index. Thus, the available multi-indices for $Q_{t_1}$
are of the form $(x, 1)$, where $x \in I_{t_2}$.

A single node's (e.g.,~associated with
subset of modes $t$) index restriction has to
hold for all successors
as we recursively visit the tree nodes in a top down
fashion; 
thus, tensors $\T{A}_1, \T{A}_2$ are passed in each \treeparam~call
so as to avoid starting the index dropping from scratch
at each  \nestedsampl~call.
Those tensors are obtained by finding the subset
of tensor entries of $\T{A}$ that obey to the rule of Relation~\ref{eq:restrict}.

\section{Experiments} \label{sec:Experiments}
\subsection{Setup}
Our experiments were conducted on a server running
the Red Hat Enterprise 6.6 OS with 64 AMD Opteron processors
($1.4$ GHz) and 512 GB of RAM. We used Matlab R2015a as the programming
framework as well as Matlab Tensor Toolbox v2.6~\cite{TTB_Software}
in order to support tensor operations.
In order to promote reproducible and
usable research, our code is \textit{open-sourced} and \textit{publicly
available}~\footnote{\url{http://www.cc.gatech.edu/~iperros3/src/sp_htucker.zip}}.

The methods under comparison are the following:
\begin{itemize}
\item {\bf Sparse H-Tucker (Sequential)}: Sparse H-Tucker implementation with sequential execution in Phase 2. 
\item {\bf Sparse H-Tucker (Parallel)}: Sparse H-Tucker implementation with parallel execution in Phase 2. 
\item {\bf H-Tucker}: Hierarchical Tucker  implementation provided in htucker toolbox~\cite{Kressner2014}\footnote{In order to enable sparse tensor input, 
we modified the computation of the left leading singular vectors by re-directing to the "nvecs" function of the "sptensor" Tensor Toolbox class.
}; 
\item {\bf CP-ALS}: Tensor Toolbox~\cite{TTB_Software} implementation;  and
\item {\bf Tucker-ALS}: Tensor Toolbox~\cite{TTB_Software} implementation of HOOI. 
\end{itemize}

%

\subsection{Experiments on real healthcare data}\label{subsec: ExpReal}
\noindent\textbf{Dataset and task description} We used \textit{publicly available}
healthcare data for our experimental
evaluation. The dataset is called MIMIC-II and can be found 
in~\cite{Saeed2011}~\footnote{http://physionet.org/mimic2/}.
It contains disease history of $29,862$ patients where an overall of $314,647$
diagnostic events are recorded over time. 
The task is about extracting co-occurring patterns
of different diagnoses in patient records, in order to better
understand the complex interactions of disease diagnoses. 
To acquire accurate phenotyping, we exploit the domain knowledge provided by the
International Classification of Diseases (ICD) hierarchy~\cite{ICD}
and guide the tensor construction with it. The ICD hierarchy consists of a collection
of trees representing hierarchical relationships between diagnoses. As such,
diagnoses belonging to the same diagnostic family reside
under the same sub-tree. We map each tensor
mode to a node of the top-level hierarchy. Thus, the order of our tensor will be equal to the number of top-level nodes. Furthermore, the lower-level diagnoses that are contained in each top-level node
will be the elements of a tensor mode.



\noindent\textbf{Input tensor construction} In order to end up with
the previously described tensor, we sum over the number of
co-occurrences for each multi-index of diagnoses.
This means that each one of the input tensor
cells contains the sum of the corresponding diagnoses
found for all patients. For example, consider the case
of a $3$-order tensor $\T{T}$ where each one of the $3$
modes corresponds to a group of similar diseases.
If a certain combination of diseases $(i, j, k)$ is co-occurring
in $3$ patients out of the whole dataset, then
$\T{T}(i, j, k) = 3$. Since the top-level of the ICD
hierarchy contains 18 nodes, our complete tensor input
is an $18$-order tensor. For the purposes of our experimental evaluation, we constructed tensors from the same dataset with varying tensor order
(less than $18$) by limiting the subset of
disease groups.
Also, to each one of the modes, we added an additional element
corresponding to "no disease", so that we model the case
when a mode does not participate in a certain co-occurrence at all.

\noindent\textbf{Cost-Accuracy Trade-offs}
At first, we would like to examine the re-construction error achieved by
the methods under comparison, as a function of the cost (time/space). 
Since many baseline methods do not scale to higher orders, we decide to use  $4$-order tensor where all methods can run without memory issues. 

\begin{figure*}[t]
	\centering
	\begin{tikzpicture}
	\begin{customlegend}[legend columns=2,legend style={draw=none,column sep=5ex},legend entries={\hspace{-5mm}Sparse H-Tucker,\hspace{-5mm}Sparse H-Tucker (Seq. Phase 2),\hspace{-5mm}H-Tucker~\cite{Kressner2014}, \hspace{-5mm}CP-ALS~\cite{TTB_Software}, \hspace{-5mm}Tucker-ALS~\cite{TTB_Software}}]
	\addlegendimage{color=mycolor1,solid,line width=1.2pt,mark size=3.5pt,mark=asterisk,mark options={solid},forget plot}
	\addlegendimage{color=mycolor5,solid,line width=1.2pt,mark size=1.2pt,mark=*,mark options={solid},forget plot}
	\addlegendimage{color=mycolor2,solid,line width=1.2pt,mark=square,mark options={solid},forget plot}
	\addlegendimage{color=mycolor3,solid,line width=1.2pt,mark size=3.5pt, mark=triangle,forget plot}
	\addlegendimage{color=mycolor4,solid,line width=1.2pt,mark=o,mark options={solid},forget plot}
	\end{customlegend}
	\end{tikzpicture}\\ \vspace{.2cm}
	\centering
	\includegraphics[scale = .33]{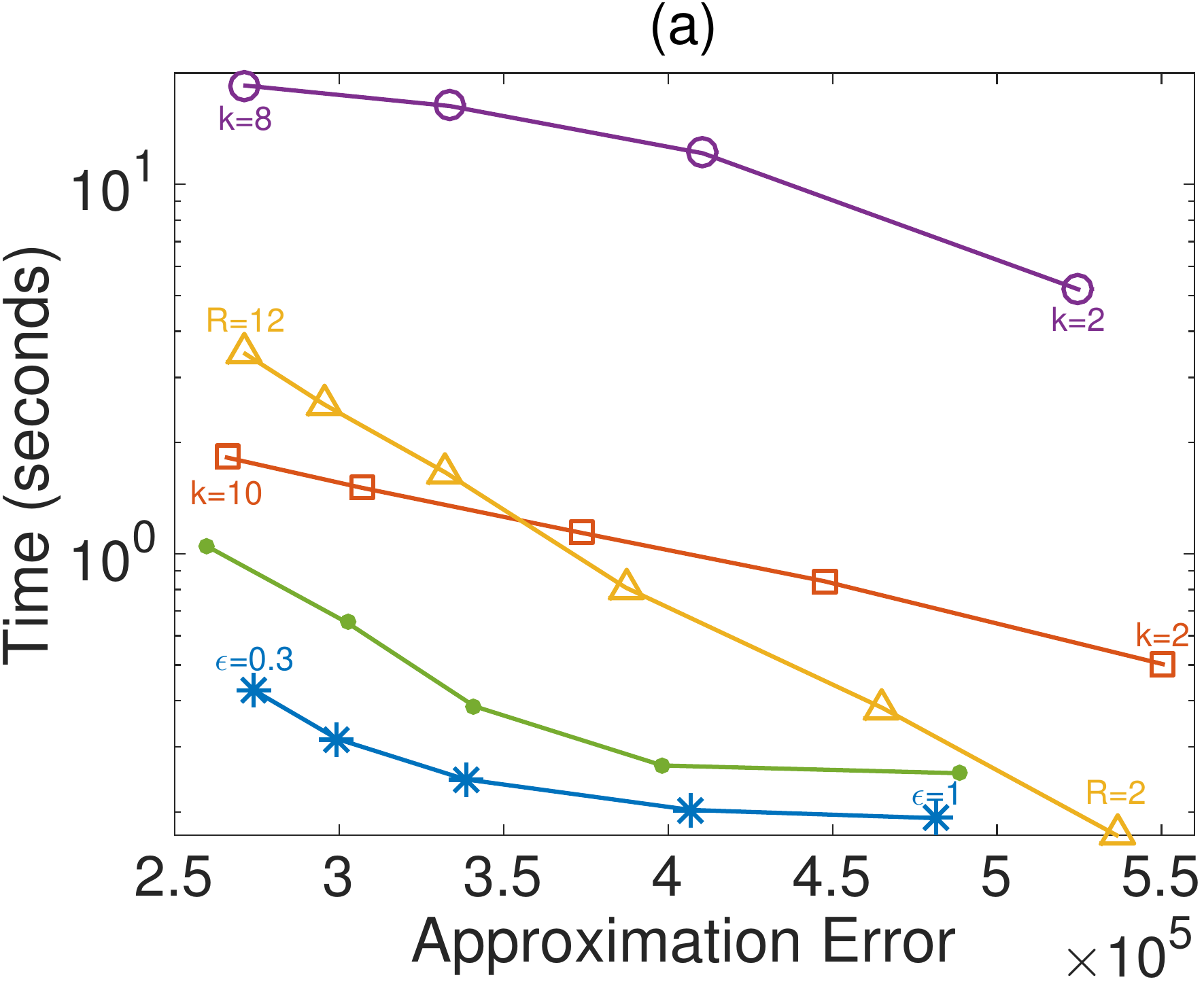}
	\includegraphics[scale = .33]{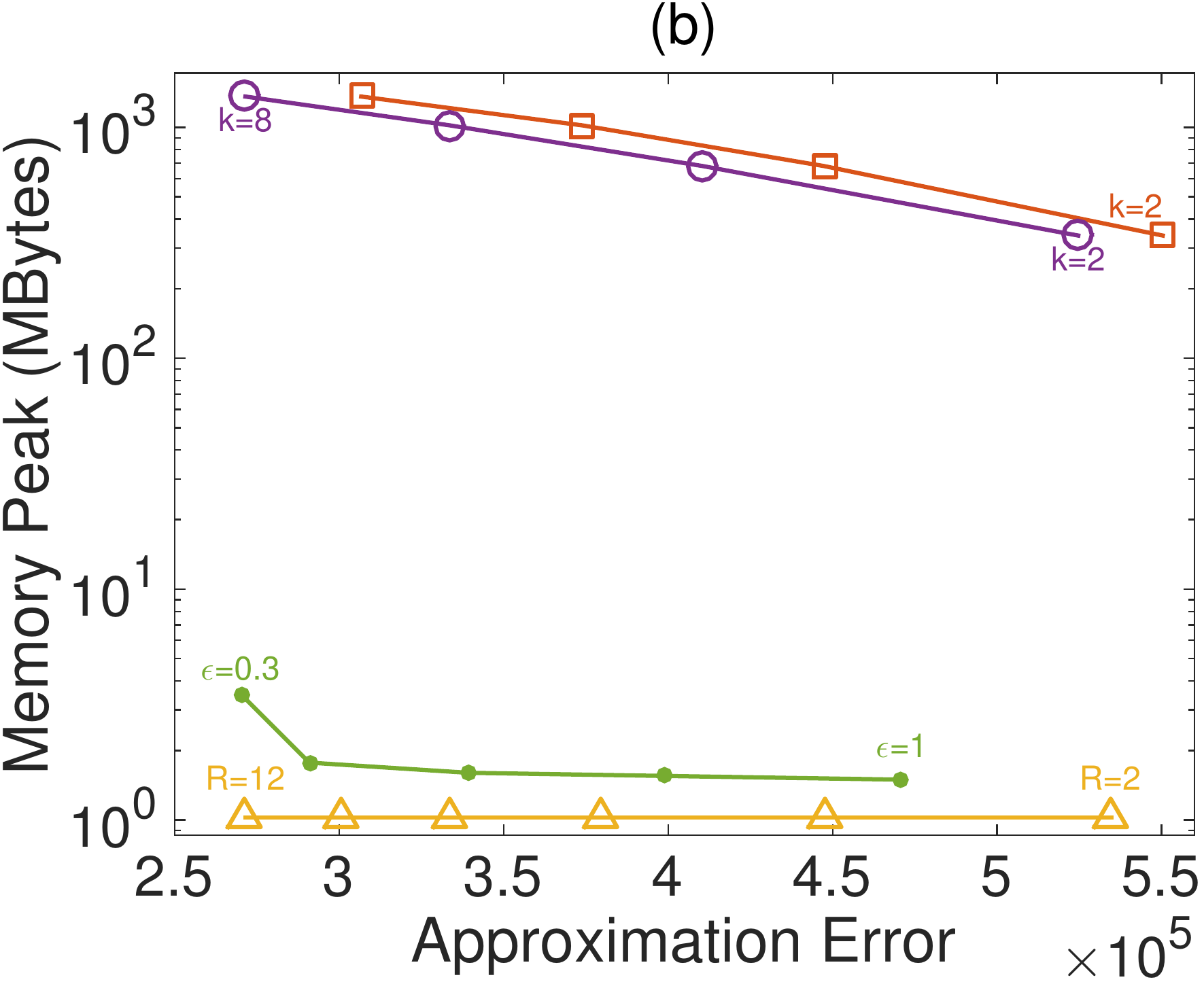}
	\caption{\small \textit{a)} Observed time and re-construction error (Frob. norm). \textit{b)} Observed peak
		space requirements and re-construction error (Frob. norm). Input tensor: \textbf{$\mathbf{4}$-order tensor} (number of non-zeros: $11100$, total size: $1.54\times10^9$) created from the \textbf{MIMIC real dataset}. The Sparse H-Tucker and H-Tucker methods run for the same
		balanced dimension tree. Points of Sparse H-Tucker correspond to runs with different $\epsilon$
		parameter: $\epsilon\in \{1, .8, .6, .4, .3\} $. Points of H-Tucker correspond to runs
		with varying low-rank parameter : $k = \{2, 4, 6, 8, 10 \}$ for
		the left and $k = \{2, 4, 6, 8\}$ for the right panel. 
		Points of CP-ALS correspond to runs with varying number of target
		rank-one factors: $R = \{2, 4, 6, 8, 10, 12 \}$.
		Points of Tucker-ALS correspond to runs with varying target
		rank: $k = \{2, 4, 6, 8\}$.}
	\label{fig:qual_err_mem}
\end{figure*}

In Figure~\ref{fig:qual_err_mem}, we observed the time-error and space-error trade-offs
for the methods under comparison.  Here we varied the
parameters governing the quality approximation in all cases.
The results were averaged over 10 runs in order to avoid fluctuations caused by random artifacts.
For our implementation executing Phase 2 in parallel, we set the number of
Matlab workers to $8$ (through the "parpool" command).

The superiority of Sparse H-Tucker as compared to all methods
in terms of the time-error trade-off is evident. In particular,
it achieves close to an order of magnitude gain ($8$x) as compared to the
CP-ALS method and $66$x
gain as compared to the Tucker-ALS. It is worth stressing out that even
with a sequential execution of Phase 2, Sparse H-Tucker outperforms
traditional tensor methods.

As concerns the space-approximation error trade-off, the savings of Sparse H-Tucker
against Tucker and H-Tucker for the
same error are remarkable: it achieves over than $2$ orders of magnitude reduction
on the peak memory allocation ($396$x). The reasons behind this stark difference
lie in the fact that Sparse H-Tucker does not form any huge, dense intermediate result
that increases with the tensor order. We were not able to reliably measure
the peak allocation memory of our method using parallelism in Phase 2. However,
we empirically noticed that the total memory required in this case is still orders of
magnitude less than the one required by Tucker and H-Tucker.

Tucker-ALS method achieves the worst time-error tradeoff, while requiring the same peak memory requirements as the H-Tucker for the same low-rank parameter. Both methods
form the same $d$-order dense tensor, either as a final output or as an intermediate result. Tucker-ALS just achieves slightly better approximation for the same space.

\begin{figure*}[t]
	\centering
	\begin{tikzpicture}
	\begin{customlegend}[legend columns=2,legend style={draw=none,column sep=5ex},legend entries={\hspace{-5mm}Sparse H-Tucker,\hspace{-5mm}Sparse H-Tucker (Seq. Phase 2),\hspace{-5mm}H-Tucker~\cite{Kressner2014}, \hspace{-5mm}CP-ALS~\cite{TTB_Software}, \hspace{-5mm}Tucker-ALS~\cite{TTB_Software}}]
	\addlegendimage{color=mycolor1,solid,line width=1.2pt,mark size=3.5pt,mark=asterisk,mark options={solid},forget plot}
	\addlegendimage{color=mycolor5,solid,line width=1.2pt,mark size=1.2pt,mark=*,mark options={solid},forget plot}
	\addlegendimage{color=mycolor2,solid,line width=1.2pt,mark=square,mark options={solid},forget plot}
    \addlegendimage{color=mycolor3,solid,line width=1.2pt,mark size=3.5pt, mark=triangle,forget plot}
	\addlegendimage{color=mycolor4,solid,line width=1.2pt,mark=o,mark options={solid},forget plot}
	\end{customlegend}
	\end{tikzpicture}\\ \vspace{.2cm}
	\centering
	\includegraphics[scale=.22]{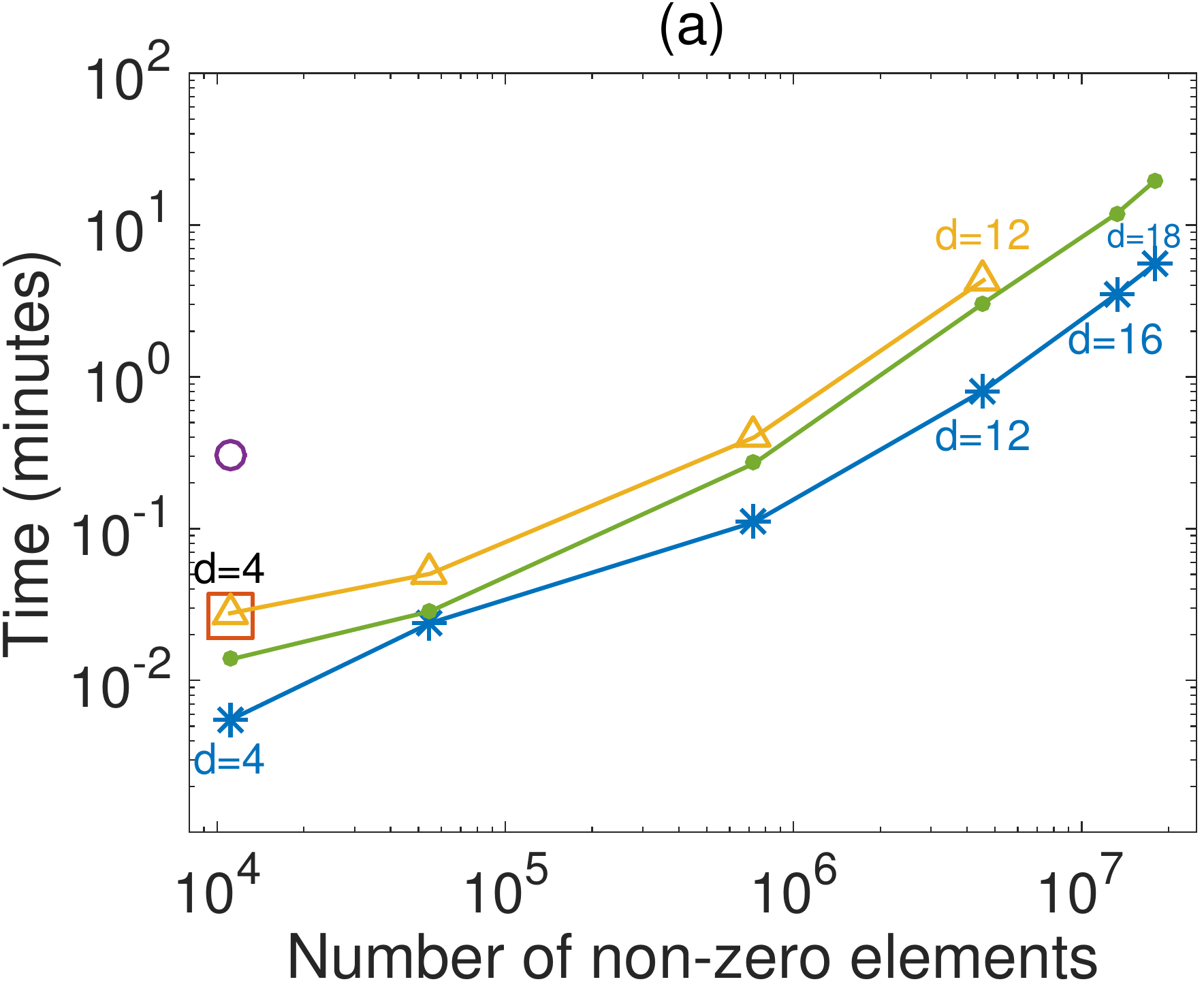}
	\includegraphics[scale=.22]{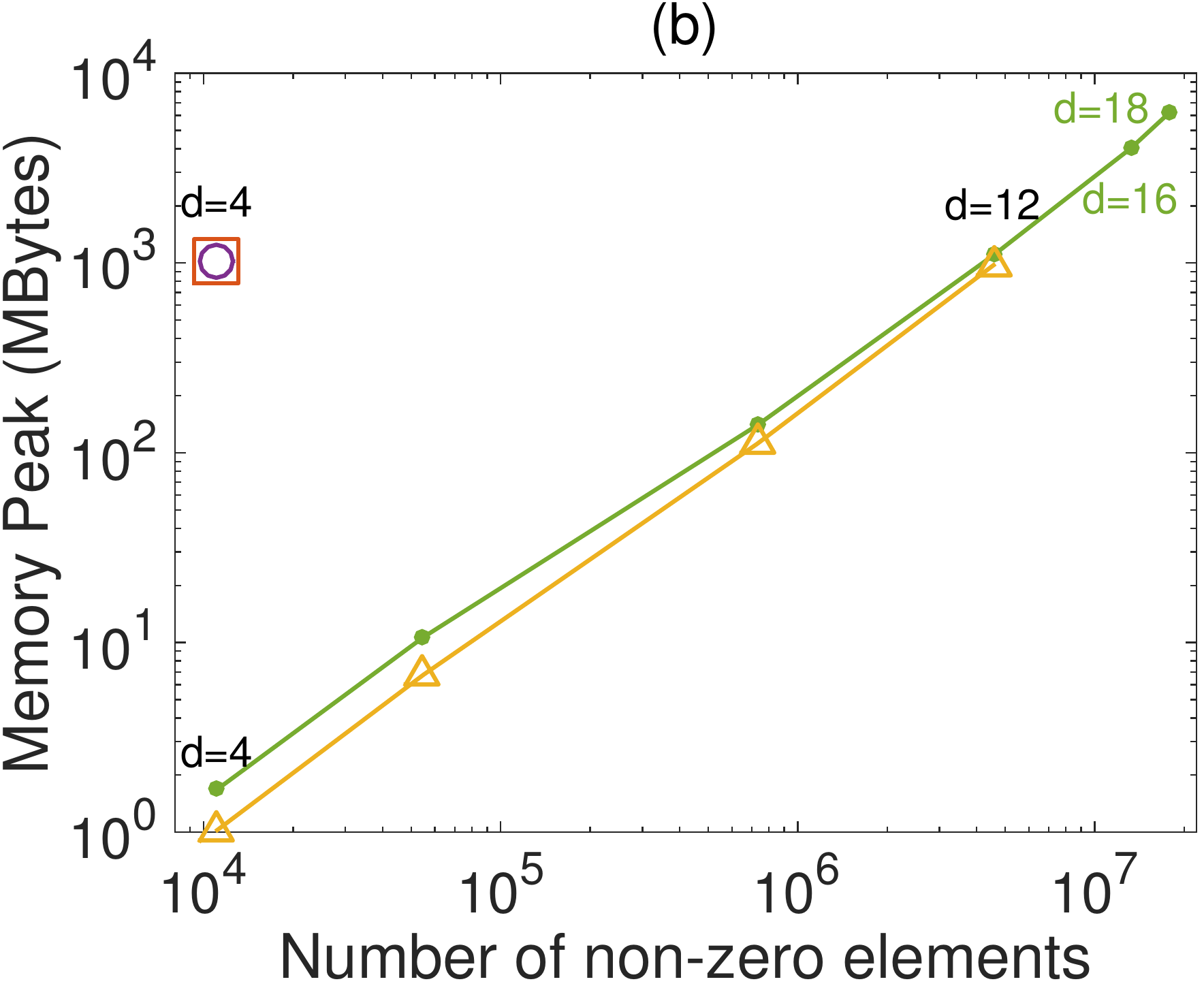}
	\includegraphics[scale=.22]{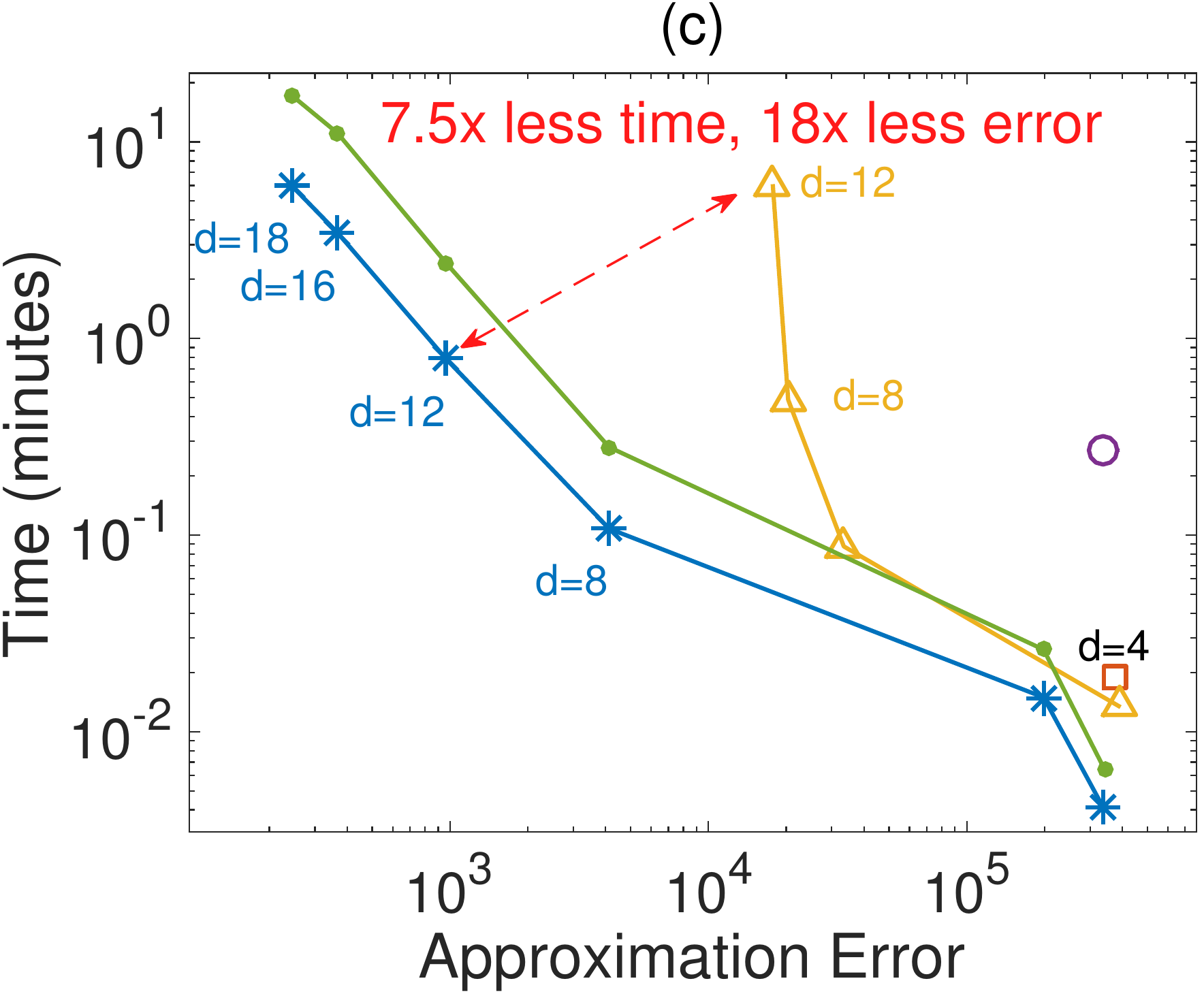}
	\caption{\small \textit{a)} Observed time and number of non-zero tensor elements \textit{b)} Observed peak space requirements and number of non-zero tensor elements. \textit{c)} Observed time and re-construction error (Frob. norm) of
		sampled non-zero entries. The methods under comparison are:
		Sparse H-Tucker ($\epsilon=0.6$), H-Tucker ($k=6$), CP-ALS ($R=6$) and Tucker-ALS ($k=6$).
		The Sparse H-Tucker and H-Tucker methods run for the same balanced dimension tree.
		Each one of the points corresponds to tensors of \textbf{increasing density and order $(d)$}
		created from the \textbf{MIMIC real dataset}. A complete description of
		the tensors used is in Table~\ref{tab: tensorstats}.}
	\label{fig:scale}
\end{figure*}

\begin{table}
	\centering
			\begin{tabular}{|c|c|c|} \hline
			\#Non-zeros (approx.) & Total size & Order \\ \hline
			$11$ K &  $1.5\times10^9$ & $4$ \\ \hline 
			$55$ K &  $10^{17}$ & $6$ \\ \hline 
			$730$ K &  $1.9\times10^{22}$ & $8$ \\ \hline 
			$4.6$ Mil & $2.1\times10^{33}$ & $12$ \\ \hline 
			$13$ Mil & $1.2\times10^{44}$ & $16$ \\ \hline 
			$18$ Mil & $4.7\times10^{49}$ & $18$ \\ \hline 
		\end{tabular}
	\label{tab: tensorstats}
	\caption{Description of tensors used in our experiments derived from real data}
\end{table}

\noindent\textbf{Scalability} We also conducted experiments in order to
assess the scalability behavior of the methods under comparison, with respect to both
the time and space required for increasing tensor order and number of non-zero
elements. The input tensors having different density and order were constructed as explained above.
For Sparse H-Tucker (parallel), we set the number of
Matlab workers to $16$, so as to exploit the full parallelism
potential of our method for higher orders. We were not able
to reliably measure the memory overhead for this version of our approach.
Still, we empirically remark that the memory required for parallel Sparse
H-Tucker shares the same scalability properties
as the sequential version.
The results are presented in Figure~\ref{fig:scale}.
It is remarkable that the H-Tucker factorization could not run for none but the $4$-order tensor case.
For the $6$-order case and beyond, the memory it required exceeded the available memory
of our server. The same behavior is observed in the case of Tucker-ALS.
On the other hand, despite having comparable scalability behavior to Sparse H-Tucker,
the CP method could not factorize the highest order tensors (16, 18) due to
numerical issues (matrix being close to singular, in a sub-problem assuming
a full-rank matrix). Our proposed Sparse H-Tucker enjoys near-linear scalability properties
with respect to increasing the non-zero elements or tensor orders for both time
and space requirements.

\noindent\textbf{Cost-Accuracy Trade-off for increasing orders} We
would finally like to evaluate the time-error trade-off as the tensor order increases.
It was intractable for any method
to re-construct the full (dense, due to approximation
errors) tensor for any order but the $4$th;
as such, we evaluated a random sample of $50$K out of all the non-zero tensor elements, for each
one of the methods (element-wise evaluation). Then, we measured the approximation
error of those re-constructed entries with the real ones from the original input tensor.
Since the $4$th order tensor contained less than $50$K non-zero values, we measured
the error for the whole tensor.
In Figure~\ref{fig:scale}c), we present the results of this experiment. We would like to
highlight the fact that as the tensor order increases, our method achieves increasingly
beneficial cost-error trade-offs over the CP-ALS method. In particular, for the $12$-order
tensor,  Sparse H-Tucker achieves \textit{$18$x \textbf{reduction} of the re-construction error
in $7.5$x \textbf{less} time}.

\subsection{Disease phenotyping case study} \label{sec:case}
In this section, we apply Sparse H-Tucker method to disease phenotyping.
The qualitative analysis refers to the results of factorizing the full $18$-order
disease co-occurrence tensor.

The factors of the Sparse H-Tucker model are fit
according to a certain tree structure. Such a tree can be obtained directly
from existing knowledge such as a domain medical ontology
or derived from data.  In this case study, we build this dimension tree in a completely data-driven fashion using hierarchical clustering.
For each one of the $m$ non-zero values of the input tensor,
we create a binary vector of size $d$, the order of the input tensor.
This vector contains ones in the non-null positions of
each specific entry.
The columns of the $m \times d$ matrix formed are considered
as data points into a $m$-dimensional space
and are hierarchically clustered according to the Jaccard coefficient.
The tree construction for the H-Tucker model
is attempted by the recent work in~\cite{Ballani2014}.
However, the cost of their algorithm is prohibitive.

\noindent\textbf{Interpretation of output factors} We
propose the following interpretation of the output
factors:
the non-zero elements that correspond to each one of
the column vectors of the matrices $(\M{U}_t)_{t \in \mathcal{L(T_I)}}$
form a concept for each individual mode $t$ of the input tensor.
Also, the numerical values of those elements are clear
indicators of their ``contribution'' to each concept, since these
are \textit{actual fibers containing co-occurrence counts} from 
the input tensor. 

As concerns the interpretation of transfer tensors $\T{B}_t$ with
$ \{ t_1, t_2 \}  = s(t)$, they should be considered as
reflecting the interactions between the concepts of the successor
nodes $t_1, t_2$. Thus, the  $(\T{B}_t)_{(i, j, v)}$ elements having the
largest absolute value within each $i$-th slice reflect
a joint concept formed through the $j$-th concept of $t_1$
and the $v$-th concept of $t_2$. Also, due to our tree construction,
the most significant concept interactions are expected
to emerge in a bottom-up fashion, which facilitates the interpretability
if one wants to focus on the dominant emerging concepts.

\begin{table}[t]
	\centering
	\scriptsize
	\begin{tabular}{|c|c|}
		\hline
		\textbf{Diagnostic family} & \textbf{Grouped clinical concepts} \\
		\hline \hline
		Endocrine, Nutritional, Metabolic & Pure hypercholesterolemia , Type II diabetes mellitus, \\
		Diseases and Immunity Disorders & Other and unspecified hyperlipidemia \\ \hline
		
		\multirow{2}*{Diseases of the Circulatory System} & Coronary atherosclerosis of native coronary artery, \\ & Hypertension, Atrial fibrillation, Congestive heart failure \\ \hline
		
		Diseases of the Blood & Anemia unspecified, Acute posthemorrhagic anemia, \\ 
		and Blood-Forming Organs & Thrombocytopenia, Secondary thrombocytopenia \\ \hline

		\multirow{4}*{Diseases of the Respiratory System} & Chronic airway obstruction, Asthma unspecified type \\ & without mention of status asthmaticus, Iatrogenic \\ & pneumothorax,  Pulmonary collapse, Pleural
		effusion, \\ & Pneumonia organism unspecified  \\ \hline
		
		Symptoms, Signs, Ill-defined conditions & Undiagnosed cardiac murmurs \\ \hline
		
		\multirow{2}*{Infectious and Parasitic Diseases} & Other Staphylococcus infection in conditions \\ & classified elsewhere and of unspecified site, Septicemia \\ \hline
	\end{tabular}
	\caption{Dominant phenotype emerging through Sparse	H-Tucker}
	\label{table:phenotyping}
\end{table}

\noindent\textbf{Qualitative analysis} We now
describe the qualitative results of our application, as they were examined
by a domain expert who verified their clinical value and meaningfulness.
Our target is to extract clinically meaningful connections
between diagnoses from different diagnostic families,
which could potentially co-occur and form valuable phenotypes.
The most significant concepts grouped together as the result of
applying our tensor factorization method, are shown in Table~\ref{table:phenotyping}.

At first, the connections within each diagnostic family reflect
well-known clinical associations. For example, concerning
intra-mode connections of the endocrine-related diseases,
inherited hypercholesterolemia is known to predispose a patient
to develop hyperlimidemia due to the inability of receptors in cells
to bind cholesterol. Also, hypercholesterolism and hyperlipidemia
are associated with type II diabetes mellitus.

The most important aspect of our results is that the inter-mode
relationships reflect meaningful disease co-occurrences as well.
The connection between elements of endocrine-related and of circulatory system diseases reflects a well-known association,
since many diabetes patients may also be hyperlipidemic.
Also, hypercholesterolemia and hypertension are known to have synergistic
effects on coronary function. 
Furthermore, the grouping of blood-related diseases with the above
is clinically meaningful, since the blood disease anemia is known to co-occur with
them. 
In addition, the coupling of the
extracted respiratory-related diseases to the aforementioned groups,
is also known to have clinical association. For example, hypercholesterolemia
is a potential risk factor for asthma 
and pre-existing heart failure may impact pneumonia development. 
The infectious diseases emerging could also form a phenotype with the
above, since staphylococcus directly affects heart valves'
functionality. 
Finally, cardiac murmurs are associated with abnormalities
contained in circulatory and blood-related diseases. 


%
%
%
%

\section{Conclusion} \label{sec:conclusion}
In this work, we propose a scalable high-order tensor factorization
method specifically designed for data analytics.
Our experiments on real healthcare data established the accuracy
and scalability of our approach.
Also, its application to the problem of disease phenotyping
confirmed its usefulness for healthcare analytics and
verified the correctness of our way of interpreting the resulting factors.
This work is the first to use the tensor networks' formalism in practice
for unsupervised learning in data mining applications. 
We would like to stress the fact that Sparse H-Tucker
is not limited to healthcare applications; healthcare is just
the focus of the current work and the application on more datasets and domains
is left as a future work. Besides this, despite being designed to tackle
high-order tensors, our proposed method is not limited to them
and obvious benefits can be seen even in the case of low-order
tensors, as we experimentally verified.
Future work will focus on further examining
the underlying tree's construction and the
method's theoretical properties.



\appendix
\section*{APPENDIX}
\textbf{Proof of Equation~\eqref{eq:B_constr}}
 \begin{proof}
The main target is to prove that if we directly form
the $\T{B}_t$ tensors through Relation~\eqref{eq:B_constr},
then Relation~\eqref{eq:U_constr} holds for interior nodes as well.
Formally, we want to prove that, if Relation~\eqref{eq:B_constr}  holds
and $\{ t_1, t_2 \} =  s(t) $ then:
\begin{equation}
( (\M{U}_t)_i )_{t \in  \mathcal{I(T_I)} } = \M{A}^{(t)}(:, q_i) , q_i \in Q_t
\label{eq:to_prove}
\end{equation}

We will prove the above proposition for the nodes
of the penultimate level of the tree. By induction, this will hold for all interior nodes.
 
Due to the restriction on each node's available
column indices w.r.t.~its parent node (Relation~\eqref{eq:restrict}), 
the nestedness property of Relation~\eqref{eq:nestedness} holds,
so that we have (element-wise):
\begin{equation}
(\M{U}_t)_{i_t, i} = \sum_{j \in Q_{t_1}} \sum_{l \in Q_{t_2}}
  (\T{B}_t)_{i, j , l} ~(\M{U}_{t_1})_{i_{t_1}, q_j}~(\M{U}_{t_2})_{i_{t_2}, q_l}
\label{eq:elemwise_nested}
\end{equation}
where $i_t \in I_t$ with $i_t = ( i_{t_1}, i_{t_2} )$ and $q_i \in Q_t$.

At this point, let the assumption that Relation~\eqref{eq:B_constr}  holds.
Then, Relation~\eqref{eq:elemwise_nested} gives:
\begin{equation}
\begin{split}
(\M{U}_t)_{i_t, i} = \sum_{j \in Q_{t_1}} \sum_{l \in Q_{t_2}}
 \sum_{p \in P_{t_1}} \sum_{q \in P_{t_2}} &
   (\M{M}_{t_1})_{q_j, p}~\M{A}^{(t)}_{\left(p,q\right), q_i}~(\M{M}_{t_2})_{q_l, q}   \\
  & (\M{U}_{t_1})_{i_{t_1}, q_j}~(\M{U}_{t_2})_{i_{t_2}, q_l}
\end{split}
\label{eq:long_elemwise}
\end{equation}

Relation~\eqref{eq:U_constr} (direct column fiber sampling
for leaf nodes) holds by construction.
Thus, Relation~\eqref{eq:long_elemwise} gives:
\begin{equation}
\begin{split}
(\M{U}_t)_{i_t, i} = \sum_{q \in P_{t_2}}  \sum_{l \in Q_{t_2}}
 \sum_{p \in P_{t_1}}  \sum_{j \in Q_{t_1}} &
   \M{A}^{(t_1)}_{i_{t_1}, q_j}~(\M{M}_{t_1})_{q_j, p}~\M{A}^{(t)}_{\left(p,q\right), q_i} \\
   & (\M{M}_{t_2})_{q_l, q}~\M{A}^{(t_2)}_{i_{t_2}, q_l}
\end{split}
\label{eq:long_elemwise2}
\end{equation}

By definition of the CUR decomposition, under the assumption that it
is exact, we have:
\[  \M{A}^{(t)}_{ \left(i_{t_1}, q \right), q_i } = \sum_{p \in P_{t_1}}  \sum_{j \in Q_{t_1}} 
   \M{A}^{(t_1)}_{i_{t_1}, q_j}~(\M{M}_{t_1})_{q_j, p}~\M{A}^{(t)}_{\left(p,q\right), q_i} \]
Thus, Relation~\eqref{eq:long_elemwise2} gives:
\begin{equation}
(\M{U}_t)_{i_t, i} = \sum_{q \in P_{t_2}}  \sum_{l \in Q_{t_2}}
    \M{A}^{(t_2)}_{i_{t_2}, q_l}~(\M{M}_{t_2})_{q_l, q}~\M{A}^{(t)}_{\left(i_{t_1}, q\right), q_i }
    = \M{A}^{(t)}_{\left(i_{t_1}, i_{t_2}\right), q_i } 
\label{eq:long_elemwise3}
\end{equation}
where the last equation follows again from the CUR decomposition.
Since we ended up to Relation~\eqref{eq:to_prove}, then using
Relation~\eqref{eq:B_constr} is correct. 

 \end{proof}


\bibliographystyle{abbrv}
\bibliography{sigproc}


\end{document}